\documentclass[runningheads]{llncs}
\usepackage[T1]{fontenc}
\usepackage{graphicx}
\usepackage{amsmath}
\usepackage{booktabs}
\usepackage{tabularx}
\usepackage{subcaption}
\usepackage{csquotes}
\usepackage{amssymb,amsfonts}
\usepackage{pifont}
\usepackage{bbding}
\usepackage{hyperref}
\usepackage[
    style=ieee,
    citestyle=numeric-comp,
    backend=biber,
    hyperref=true,
    maxnames=2,
    maxbibnames =3,
    minnames=2,
    mincrossrefs=1,
    natbib=true,
    sorting=none,
    sortcites=true,
    url=false,
    doi=false,
    eprint=false
]{biblatex}
\AtEveryBibitem{\clearname{translator}\clearlist{publisher}\clearfield{pagetotal}\clearfield{pagetotal}\clearfield{issn}\clearfield{editor}\clearfield{month}\clearfield{notes}\clearfield{language}}
\DeclareFieldFormat
  [article,inbook,incollection,inproceedings,patent,thesis,unpublished]
  {title}{{#1}}

 \def\BibTeX{{\rm B\kern-.05em{\sc i\kern-.025em b}\kern-.08em
    T\kern-.1667em\lower.7ex\hbox{E}\kern-.125emX}}
\AtBeginBibliography{\footnotesize}

\usepackage[skip=4pt]{caption}   
\graphicspath{{.}}   

\begin{document}

\title{Are ECG Foundation Models Always Better? A Study of Heart Failure Risk Prediction and Fairness}

\author{Tanay Patel\inst{1}(\Envelope)\orcidID{0009-0004-7287-9824}  \and
Xuelong An \inst{1}\orcidID{0009-0006-3235-4059} \and
Chen Chen~(\Envelope)\inst{1}\orcidID{0000-0002-3525-9755
}}
\authorrunning{T. Patel et al.}
\institute{School of Computer Science, University of Sheffield, Sheffield, UK\\
\email{
\{tvpatel1, chen.chen2\}@sheffield.ac.uk}}
\titlerunning{Accuracy and Fairness in ECG Foundation Models}
\maketitle

\begin{abstract}

ECG foundation models which leverage self-supervised learning (SSL), can learn data-efficient, transferable representations from large-scale unlabelled ECG recordings to enable strong performance across a wide range of cardiac disease diagnostic tasks. However, few studies have explored the benefits of using foundation models for risk prediction tasks or whether they can improve demographic equity in these more challenging, yet clinically important, settings. To our knowledge, we present the first race-stratified fairness evaluation of a self-supervised ECG foundation model for incident heart failure (HF) risk prediction. We compare a supervised ResNet1D with an ECG foundation model (ECG-FM) using the Harvard–Emory ECG Database, with predictive performance measured by the concordance index (C-index) and fairness assessed across sex and racial subgroups using the Performance Gap (PG) and Coefficient of Variation (CV). Surprisingly, ResNet1D achieved the highest predictive performance, with a mean C-index of 0.800, significantly outperforming the pretrained ECG-FM (0.791) and the randomly initialised ECG-FM (0.789) ($p<0.05$). In contrast, the pretrained ECG-FM yielded the smallest racial disparities, while the randomly initialised ECG-FM exhibited the largest, indicating that pretraining could reduce performance disparities across demographic subgroups. Sex-related disparities were very mild across all models, likely because the dataset had a balanced sex distribution. Our study demonstrates that demographic fairness in clinical risk prediction cannot be characterised by aggregate predictive performance alone. As racial imbalance remains common in clinical datasets, algorithmic innovations should be accompanied by routine demographic subgroup evaluation to ensure the fairness, reliability, and trustworthiness of AI-ECG models before clinical deployment.

\end{abstract}
 
\section{Introduction}
Heart Failure (HF) is a significant condition in which the heart cannot pump sufficient blood to meet the body's metabolic demands, affecting 55--64 million people worldwide with a rising global mortality rate~\cite{valente_global_epidemiology_2026}. Early detection and prevention of HF could significantly reduce its impact. 
Recent advances in artificial intelligence (AI) have demonstrated promise in preventive cardiac care, e.g., utilising low-cost electrocardiogram (ECG) to support early cardiovascular disease diagnosis, prognosis and risk prediction~\cite{muzammil_artificial_intelligence-enhanced_2024,chen_llm}. In particular, advanced AI-ECG foundation models~\cite{ecg-fm, li_ecg_foundation_2025}, such as ECG-FM~\cite{ecg-fm} have demonstrated high diagnostic accuracy, thanks to large-scale self-supervised learning (SSL)-based pretraining on extensive unlabelled ECG corpora~\cite{nolin_lapalme_deepssl_2025}. The learned representations can transfer effectively to downstream clinical tasks and have shown to outperform conventional supervised models without pretraining, especially in settings where annotated training data for the downstream task is limited~\cite{nolin_lapalme_deepssl_2025}. 

 
Yet most evaluations focus primarily on aggregate  performance, which is insufficient for assessing the fairness and reliability of AI-ECG systems. While some studies have examined fairness in AI-ECG models, their analyses have largely been limited to classification tasks~\cite{kaur_race_2023, nolin_lapalme_deepssl_2025} and primarily focus on reporting disparities across age and sex groups. In contrast, in this work, we conduct a race-stratified fairness audit, focusing on incident heart failure (HF) survival prediction, an important clinical task. We compare a self-supervised pretrained ECG foundation model with conventional supervised baselines to assess performance across sex and racial groups. Our analysis aims to investigate whether foundation models provide more equitable risk predictions.

\section{Methodology}
\textbf{Data}
A high-risk cohort of patients with hypertension and/or prior myocardial infarction at initial screening was selected from the public, ethics-approved \href{https://bdsp.io/content/heedb/5.0/}{Harvard-Emory ECG Database (HEEDB)}~\cite{koscova_harvard_emory_2024} to study the risk of incident HF, yielding $N = 187{,}694$ patients (51.4\% Male vs.\ 48.6\% Female) with 12-lead ECG recordings, following the data selection criteria in~\cite{chen_llm}. Over a median follow-up of 30.9 months, 20{,}029 events (10.67\%) were recorded in a cohort comprising 77.9\% White (10.38\% HF), 9.8\% Black (16.35\%), and 3.0\% Asian (7.04\%) patients. The substantial racial imbalances in both cohort composition and HF incidence highlight the need for a race-stratified fairness analysis. 

\begin{figure}[htbp!]
    \centering
    \includegraphics[width=0.7\textwidth]{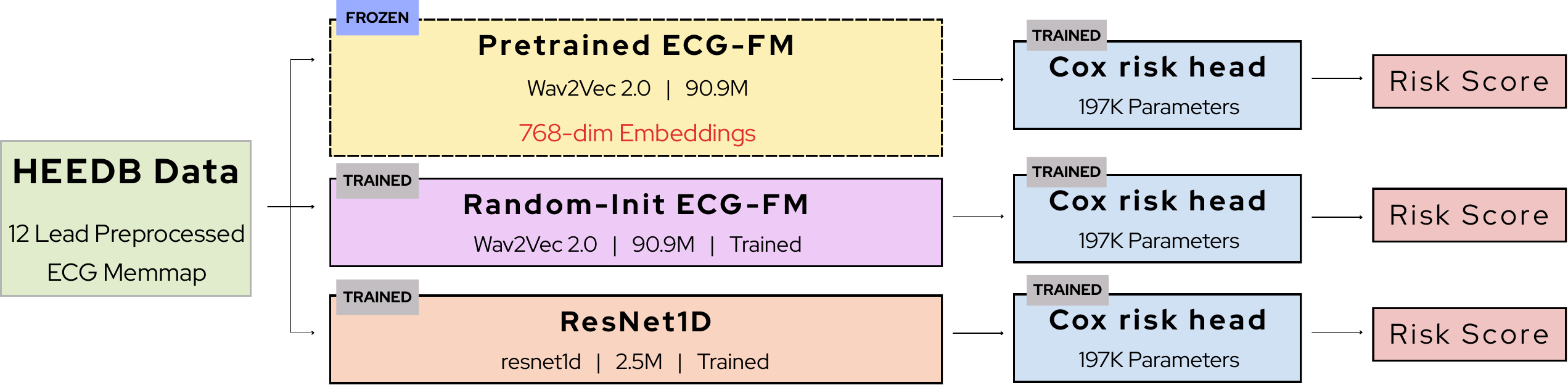}
    \caption{\textbf{Overview of Study Design.} Each ECG was resampled to 500\,Hz, band-pass filtered with baseline-wander correction, $z$-score normalised per lead, and padded or centre-cropped before being sent to the network.}
    \label{fig:experiment_pipeline}
\end{figure}

\noindent\textbf{Model Selection} 
As shown in Fig.~\ref{fig:experiment_pipeline}, three models were compared: \textbf{a) Pretrained ECG foundation model: ECG-FM}, one of the most representative ECG foundation models, pretrained from $\sim$1.5\,M unlabelled ECGs~\cite{ecg-fm} using SSL, was kept frozen and evaluated as a linear probe, exposing the prognostic utility of the pretrained representations; 
\textbf{b) Random-Init ECG-FM} trained directly from scratch without SSL pretraining for ablation study; 
\textbf{c) ResNet1D}~\cite{hong2020holmes_resnet}, one of the most popular convolutional network for ECG analysis, trained directly from scratch, which has demonstrated top-performing results in the PhysioNet Challenge. 
For models trained from scratch, random initialisation is applied. Each model receives a 12-lead ECG as input, which is processed by either a pretrained or randomly initialised encoder from the above, followed by a shared two-layer projection head ($768\!\to\!256\!\to\!1$) for risk prediction. Models were optimised using the same Cox negative log partial-likelihood loss following~\cite{chen_llm}.

\noindent\textbf{Evaluation and Fairness Metrics} 
Models were evaluated under stratified 10-fold cross-validation. We use the \textbf{C-index} to evaluate the predictive performance of risk prediction. A C-index of 0.5 corresponds to random guessing, whereas a value of 1.0 indicates perfect prediction. To evaluate fairness, we report two complementary metrics. Performance Gap (PG) quantifies the largest difference in C-index between any two subgroups within a demographic axis (sex or race)~\cite{liu_equitable_survival_2025}, while the Coefficient of Variation (CV) measures the relative variability of subgroup C-indices \cite{coefficient_of_variance}. Lower PG and CV values indicate more equitable performance across demographic subgroups.

\section{Results}

\textbf{Aggregate Performance: ResNet1D Outperforms ECG-FM in HF Risk Prediction}
Surprisingly, the highest aggregate performance is achieved using the traditional supervised ResNet1D, yielding the highest mean C-index of 0.800, which was statistically significant ($p<0.05$; Holm-Bonferroni corrected) than both Pretrained (0.791) and Random-Init ECG-FM (0.789).

\begin{figure}[t]
    \centering
    \includegraphics[width=\linewidth]{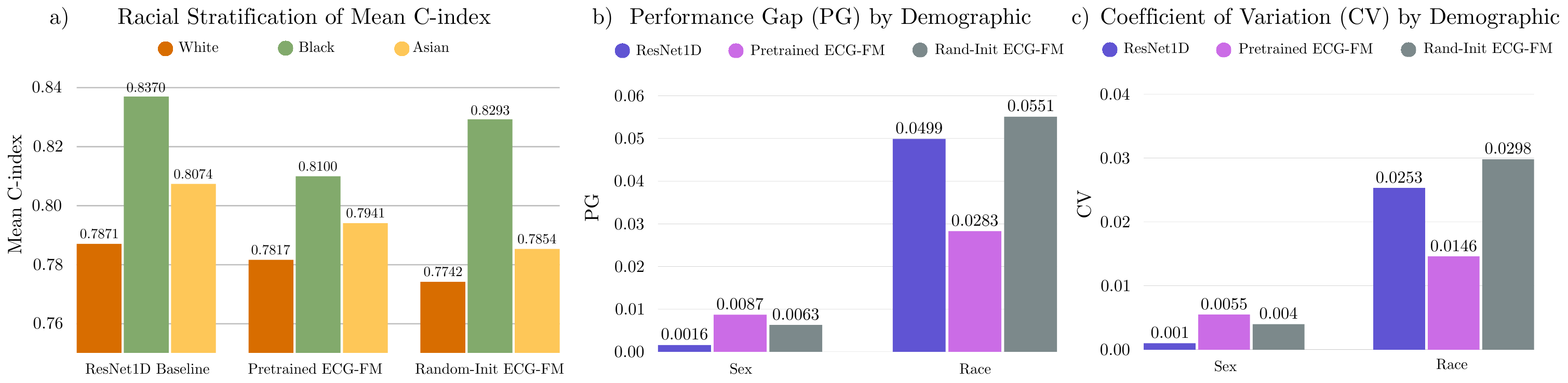}
    \caption{a) Mean C-index per racial subgroup (White, Black, Asian) over 10-fold cross-validation; b) PG and c) CV across the \textbf{Sex} and \textbf{Race} axes.}
    \label{fig:results_combined}
\end{figure}

\noindent\textbf{Fairness Evaluation: Pretrained ECG-FM Reduces Racial Disparities Despite Slightly Lower Accuracy}
We further compared the three models' performance across different racial groups, as shown in Fig.~\ref{fig:results_combined} a). It is evident that the ResNet1D approach outperforms the two ECG-FM variants across all three groups. Notably, all three models achieved relatively higher predictive performance in the Black subgroup. This may be due to the higher HF event rate in Black patients (16.35\% vs.\ 10.38\% in White patients and 7.04\% in Asian patients), which yields more admissible concordance pairs to inform model training.

Yet, as shown in Fig.~\ref{fig:results_combined}b) and~\ref{fig:results_combined}c), while all models demonstrated substantially smaller disparities across sex (PG values below 0.009 and CV values below 0.006), ResNet1D, despite its superior aggregate performance, exhibited the higher racial disparity compared to the pretrained ECG-FM, with a race Performance Gap (PG) of 0.0499 and a race Coefficient of Variation (CV) of 0.0253. In contrast, the pretrained ECG-FM achieved the lowest racial disparity (PG = 0.0283, CV = 0.0146), indicating the most consistent performance across racial groups. The randomly initialised ECG-FM, included as an ablation baseline, exhibited both lower predictive performance and the largest racial disparities (PG = 0.0551, CV = 0.0298), indicating that the fairness improvement observed in the pretrained ECG-FM is unlikely to arise from the architecture alone. Instead, the pretrained representations with SSL appear to contribute to more consistent performance across racial subgroups.

\enlargethispage{\baselineskip} 
\section{Discussion and Conclusion}

This study presents, to our knowledge, the first sex- and race-stratified fairness evaluation of a self-supervised ECG foundation model for time-to-event HF risk prediction, extending previous work that has focused on age and sex~\cite{nolin_lapalme_deepssl_2025} or simple supervised disease classification tasks~\cite{kaur_race_2023}. Our results reveal a trade-off between predictive discrimination and demographic equity. The ResNet1D achieved the strongest aggregate performance, whereas the pretrained ECG-FM, benefiting from SSL-based pretraining, yielded more equitable performance across racial subgroups at the expense of absolute predictive discrimination.

The supervised ResNet1D outperformed both the pretrained and randomly initialised ECG-FM models in aggregate HF risk prediction. One possible explanation is that the relatively low event rate and the amount of labelled training data favour a lightweight architecture that can more effectively learn task-specific features. In our study, ResNet1D contains approximately 2.5\,M parameters, compared with 90.9\,M parameters for ECG-FM. The substantially smaller model may therefore be easier to optimise and less susceptible to overparameterisation under limited supervision. In contrast, ECG-FM is designed to learn general ECG representations through large-scale pretraining. Although these representations are transferable across diverse downstream tasks, they may not be fully optimised for long-term HF risk prediction and may require more extensive data during fine-tuning to achieve their full potential.

While the supervised ResNet1D achieved the strongest discrimination within each racial group, the Pretrained ECG-FM substantially reduced racial disparities. The architecture-matched randomly initialised ablation provides directional evidence that this equity behaviour is associated with self-supervised pretraining rather than architecture alone.

Several limitations should be acknowledged. The evaluation was conducted on a single high-risk cohort, the relatively small Asian subgroup introduced greater uncertainty into its per-group estimates, and the frozen-versus-end-to-end comparison cannot fully separate the effects of pretraining from those of the training regime. Future work should consider additional foundation models and evaluate fully fine-tuned foundation models as well, formally test the statistical significance of disparity measures, and validate these findings across external cohorts. Overall, our results demonstrate that demographic fairness in clinical risk prediction cannot be characterised by aggregate performance alone. As racial imbalance remains common in clinical datasets, algorithmic innovations should be accompanied by routine demographic subgroup evaluation to ensure the fairness, reliability, and trustworthiness of AI-ECG models before clinical deployment.

\begin{credits}
\subsubsection{\ackname}
This project has received ethics approval (No. 71394) from the University of Sheffield (UoS) Research Ethics Committee. We acknowledge IT Services at UoS for the provision of services for High Performance Computing.
\end{credits}

\printbibliography

\end{document}